\documentclass[conference]{IEEEtran}
\IEEEoverridecommandlockouts
\usepackage{cite}
\usepackage{amsmath,amssymb,amsfonts}
\usepackage{algorithmic}
\usepackage{graphicx}
\usepackage{textcomp}
\usepackage{xcolor}
\def\BibTeX{{\rm B\kern-.05em{\sc i\kern-.025em b}\kern-.08em
    T\kern-.1667em\lower.7ex\hbox{E}\kern-.125emX}}
\usepackage{tabularx}
\usepackage{booktabs}
\usepackage{float}

\begin{document}
\title{Assessing ML Classification Algorithms and NLP Techniques for Depression Detection: An Experimental Case Study} 

\author{
    \IEEEauthorblockN{Giuliano Lorenzoni, Cristina Tavares, Nathalia Nascimento, Paulo Alencar, Donald Cowan}
    
    \IEEEauthorblockA{David R. Cheriton School of Computer Science\\
    University of Waterloo (UW)\\
    Waterloo, Canada\\}

    \IEEEauthorblockA{
$\left\{\textsuperscript{}glorenzo, cristina.tavares, nmoraesd, palencar, dcowan\right\}$@uwaterloo.ca 
    }
}

\maketitle

\begin{abstract}
Depression has affected millions of people worldwide and has become one of the most common mental disorders. Early mental disorder detection can reduce costs for public health agencies and prevent other major comorbidities. Additionally, the shortage of specialized personnel is very concerning since Depression diagnosis is highly dependent on expert professionals and is time-consuming. 

Recent research has evidenced that machine learning (ML) and Natural Language Processing (NLP) tools and techniques have significantly benefited the diagnosis of depression. However, there are still several challenges in the assessment of depression detection approaches in which other conditions such as post-traumatic stress disorder (PTSD) are present. These challenges include assessing alternatives in terms of data cleaning and pre-processing techniques, feature selection, and appropriate ML classification algorithms. This paper tackels such an assessment based on a case study that compares different ML classifiers, specifically in terms of data cleaning and pre-processing, feature selection, parameter setting, and model choices.
The case study is based on the Distress Analysis Interview Corpus - Wizard-of-Oz (DAIC-WOZ) dataset, which is designed to support the diagnosis of mental disorders such as depression, anxiety, and PTSD. Besides the assessment of alternative techniques,  we were able to build models with accuracy levels around 84\% with Random Forest and XGBoost models, which is significantly higher than the results from the comparable literature which presented the level of accuracy of 72\% from the SVM model. 




\end{abstract}


\section{Introduction}
Depression has affected million of people worldwide. It is considered a common mental disorder not only because of the increasing number of cases but also because of its intrinsic relation to other psychiatric and physical illnesses such as stroke, heart disease, and chronic fatigue. 
The effects of the pandemic on general mental health, the recent rise in cases of mental health issues, and the shortage of professionals specialized in the diagnosis and treatment of mental disorders such as depression all characterize a serious issue that can have several negative implications for society. 

In this context, one way to help to cope with the effects of the rising number of depression cases and the shortage of medical specialists is to introduce early detection methods. 
Recent research has evidenced that machine learning (ML) and Natural Language Processing (NLP) tools and techniques can significantly benefit the early diagnosis of depression. However, there are still several challenges in the assessment of depression detection approaches in which other conditions such as post-traumatic stress disorder (PTSD) are present. These challenges include assessing alternatives in terms of data cleaning and pre-processing techniques, feature selection, and appropriate ML classification algorithms.

This paper tackels such an assessment based on a case study that compares different ML classifiers, specifically in terms of data cleaning and pre-processing, feature selection, parameter setting, and model selection. The case study is based on the Distress Analysis Interview Corpus - Wizard-of-Oz (DAIC-WOZ) dataset, which is designed to support the diagnosis of mental disorders such as depression, anxiety, and PTSD. Besides the assessment of alternative techniques,  we were able to build models with accuracy levels around 84\% with Random Forest and XGBoost models, which is significantly higher than the results from the comparable literature which presented the level of accuracy of 72\% from the SVM model. 


This paper is structured as follows. Section II presents related work. Section III describes our experiment design and Section IV presents the results of our study. Section V provides some discussion on insights about the models, selected features, and other factors and their dependencies. Section VI describes threats to validity, and, finally, Section VII presents our conclusions and future work.  

\section{Related Work} \label{sec:related}

Regarding depression detection in general, recent studies have shown that machine learning tools can be applied to predict mental health diseases. Saqib et al. conducted a literature review to investigate the use of machine learning in the prediction of postpartum depression (PPD) and found 14 research works in this theme. Findings revealed progressive results in performance of machine learning methods and techniques that could benefit early PPD detection significantly \cite{3saqib2021machine}. 
A meta-analysis and systematic review conducted by Lee et al. examined the use of machine learning algorithms to predict mood disorders and support on the selection for treatment and therapy. Authors provided an overview of machine learning applications applied to population with mood disorders. The study indicated high accuracy results in the the prediction of therapeutic outcomes, and conclusions suggested that machine learning techniques could be powerful tools in supporting the research of mood disorders and analysis of multiple types of data from diverse sources \cite{4lee2018applications}. 

Spacic and Goran have conducted a study to investigate the use and application of text data on clinical practice using machine learning and NLP techniques \cite{5spasic2020clinical} 
The study was based on a systematic literature review aiming to collect information about research related these knowledge areas. Findings revealed major barriers to use machine learning for clinical practice such as a lack of use of large datasets to train models and difficulties in the annotation process. The authors’ conclusions suggested further research in unsupervised learning or data augmentation and transfer learning.
Finally, although we have identified related literature reviews and practical applications, there is still a need to advance the application of techniques to support depression detection models and improve their accuracy.


In terms of the ways NLP techniques can be applied in the study of clinical depression, for example, using social media data or other forms of text data,
the related efforts falls into the following categories:
\begin{itemize}
    \item The articles which compare different techniques in order to find the best model according to some evaluation metrics, such as \cite{12ganguli2020survey,13malviya2021transformers,14nandanwar2021depression,15yuki2020detecting,16tadesse2019detection,17azam2021identifying,18hossain2021social,19skaik2020using,20alghamdi2020predicting}. 
    \item The articles that proposed new detection or prediction models, platforms, or systems, namely \cite{20alghamdi2020predicting,21rastogi2018multi, 22zhang2021depa,23maniar2021depression,24zogan2021depressionnet,25jain2020mental,26trotzek2018utilizing,27uddin2022deep,28wongkoblap2021deep} 
    or articles regarding specific applications \cite{7alabdulkreem2021prediction, 22zhang2021depa,29wang2020multimodal,30mulay2020automatic,31shickel2020automatic,32sanyal2021study,33musleh2022twitter,34low2020natural,35nguyen2014affective,36zaghouani2018large}. 
\end{itemize}

 We have also found several articles which belong to more than one group. For instance, we have found articles proposing a new model and testing it along with other techniques, such as \cite{37jagtap2021use} 
 whose authors proposed a system to detect depression using Natural Language Processing (NLP) techniques and compared the performance of this method with multiple machine learning algorithms in order to identify best-performing configuration. 
 
 Similarly, article  \cite{31shickel2020automatic} 
 presents a machine learning framework for the automatic detection and classification of 15 common cognitive distortions (defined by the authors as automatic and self-reinforcing irrational thought patterns) in two novel mental health free datasets collected from both crowdsourcing and a real-world online therapy program, comparing the results of such framework with multiple traditional machine learning models/techniques (e.g XG-Boost, SVM, CNN, and RNN). Likewise, the authors of \cite{38hassan2020recognizing} 
 present an automated conversational platform that was used as a preliminary method of identifying depression associated risks and have its results compared with the ones provided by the classic SVM classifier. Moreover, this article goes further on the classification task and looks for the identification of different levels of states (happy’, ‘neutral’, ‘depressive’ and ‘suicidal’) like the work of the authors of \cite{30mulay2020automatic} which classifies different degrees (or categories) of depression (Minimal, Mild, Moderate, Severe). There are also articles on specific applications involving the proposal of new detection models, systems, or platforms in a specific context. For example, in \cite{22zhang2021depa}, the authors present DEPA, which is a self-supervised, pre-trained depression audio embedding method for depression detection, and also explore self-supervised learning in a specific task within audio processing. 



Regarding ML models and techniques applied in depression studies, two main groups of studies emerge:

\begin{itemize}
    \item First group: Articles whose techniques were applied in studies comparing traditional machine learning models to identify the best model according to some evaluation metrics (i.e. accuracy), such as in \cite{12ganguli2020survey,13malviya2021transformers,14nandanwar2021depression,15yuki2020detecting,16tadesse2019detection,17azam2021identifying,18hossain2021social,19skaik2020using}.
    \item Second group: Articles proposing new techniques for depression detection in a specific context   \cite{20alghamdi2020predicting,21rastogi2018multi,22zhang2021depa,23maniar2021depression,24zogan2021depressionnet,25jain2020mental, 29wang2020multimodal,30mulay2020automatic,31shickel2020automatic,32sanyal2021study,40haque2020transformer}. 
    These techniques were compared with the traditional models.
\end{itemize}

As for the techniques mentioned in the articles from the  first group, we can highlight: (a) Support Vector Machine (SVM) \cite{12ganguli2020survey,13malviya2021transformers,14nandanwar2021depression,15yuki2020detecting,16tadesse2019detection,17azam2021identifying,18hossain2021social,19skaik2020using}: 
(b) Random Forest (RF) \cite{12ganguli2020survey,13malviya2021transformers,14nandanwar2021depression,15yuki2020detecting,17azam2021identifying,18hossain2021social,19skaik2020using}: 
(c) Gradient Boosted Trees (XGBoost) \cite{13malviya2021transformers,19skaik2020using}: 
(d) Naive Byes Classifiier \cite{13malviya2021transformers,14nandanwar2021depression,18hossain2021social}: 
(e) Logistic Regression (LR) \cite{12ganguli2020survey,13malviya2021transformers,18hossain2021social,19skaik2020using}: 
(f) Decision Tree (DT) \cite{12ganguli2020survey,13malviya2021transformers,15yuki2020detecting,18hossain2021social,19skaik2020using}: 
(g) Extra Trees Classifier \cite{14nandanwar2021depression}: 
h) Stochastic Gradient Boost Classifier \cite{14nandanwar2021depression}:  
(i) Long short-term memory (LSTM)/ recurrent neural network (RNN) \cite{15yuki2020detecting}: 
(j) Ensemble methods \cite{15yuki2020detecting}: 
(k) Multilayer Perceptron Classifier (MLP) \cite{16tadesse2019detection}:  
(l) Multinomial Naive Bayes (NB) \cite{18hossain2021social}: 
and (m) K-Nearest Neighbor (KNN) \cite{18hossain2021social}. 

Along with most of the traditional machine learning models mentioned in the articles from the  first group, it is worth mentioning the following techniques used in the articles in the second group: a) Deep neural network classification model - Multimodal Feature Fusion Network (MFFN) \cite{20alghamdi2020predicting}: 
b) Bidirectional Encoder Representations of Transformers (BERT); and c) Robustly optimized BERT approach (RoBERTa).
Still, regarding the second group, we have identified the use of some transformer models \cite{40haque2020transformer}, 
which are semi-supervised machine learning techniques that are primarily used with text data, and constitute an alternative to recurrent neural networks in natural language processing tasks. Basically, it consists in deep learning models that adopt the mechanism of self-attention, deferentially weighting the significance of each part of the input data. It is used primarily in the field of natural language processing (NLP) and in computer vision (CV). 

In summary, although we also found some unsupervised models among the articles in the second group, most of the work in the literature seems to focus on supervised approaches such as Support Vector Machines, Random Forest, and XGBoost. Finally, Despite these advancements, further research is required to refine depression detection models and enhance their precision.


Specifically regarding the articles using the DAIC-WOZ database 
the detection of depression through machine learning (ML) and natural language processing (NLP) has presented varied approaches, with a significant body of literature exploring this domain. The methodologies range from text analysis on social media \cite{liu2022detecting} to speech emotion analysis \cite{bhavya2022speech}, highlighting the multidisciplinary nature of the field.

Text-based studies, like those conducted by Liu et al. \cite{liu2022detecting} and Chaves et al. \cite{chaves2022automatic}, focus on detecting depressive symptoms using sentiment analysis and summarization techniques on social media platforms. These studies emphasize the potential of text mining in identifying depression but also recognize the challenges of data bias and privacy concerns. In contrast, speech-based analyses, as can be found in the works of Bhavya et al. \cite{bhavya2022speech}, Squires et al. \cite{squires2023deep}, and Cummins et al.  \cite{cummins2018speech}, investigate acoustic features and speech patterns using datasets like DAIC-WOZ for emotion recognition, presenting the challenge of data scarcity and the need for robust and generalizable models.

Studies like Krishna and Anju \cite{krishna2020different} and Nguyen et al. \cite{nguyen2023multimodal} extend the exploration into multimodal domains, arguing for the combination of audiovisual cues with linguistic analysis. They highlight the potential of deep learning models to integrate these diverse data streams for higher diagnostic accuracy. This trend is echoed by Nouman et al. \cite{nouman2021recent}, which explores contactless sensing technologies, underscoring the importance of multifaceted data in enhancing predictive models.

The gaps in the literature are prominently presented, with a recurring theme being the scarcity of large and diversified data sets, a challenge that Silva et al. \cite{da2022review} and Yan et al.  \cite{yan2022challenges} address by calling for open-access resources to bolster research efforts, tackling everything from data scarcity to ethical dilemmas. More specifically, these studies offer comprehensive reviews, the former focusing on computational methods and databases and the latter on the broader challenges that AI faces in recognizing mental disorders (for example, addressing logical fallacies and diagnostic criteria that often hinder AI's potential in mental health). These reviews also highlight the gaps in standardization and the need for longitudinal analyses to advance the field. 

Similarly, Song et al. \cite{song2018human} and Janardhan and Nandshini \cite{janardhan2022improving} emphasize the role of deep learning and feature selection in improving the accuracy of depression prediction, pointing out the limitations of small data sets and advocating for larger and more diverse data collections.

Additionally some of the mentioned studies (such as  \cite{da2022review} and \cite{yan2022challenges}) methodologically reveal a preference for ensemble and hybrid ML models. This information is also evidenced in the works of Mao et al. \cite{mao2023systematic} and Janardhan and Nandhini \cite{janardhan2022improving}, which prioritize the combination of classifiers for improved predictive outcomes. In fact, Mao et al. \cite{mao2023systematic} provides a systematic review of automated clinical depression diagnosis, synthesizing findings across 264 studies. The review emphasizes the importance of acoustic features and their correlation with observable symptoms, while also pointing out methodological variations and ethical considerations that need to be addressed in future research.

The research also points to a growing dependence on sophisticated data preprocessing and feature extraction techniques, as detailed in Aleem et al. \cite{aleem2022machine} and Nouman et al. \cite{nouman2021recent}, to mitigate the problems of class imbalance and data scarcity. Indeed, Aleem et al. \cite{aleem2022machine} and Ndaba et al. \cite{ndaba2023review} discuss the complexities of data handling techniques for depression prediction, addressing the problem of class imbalance prevalent in health records. They suggest that future work should explore under-sampling techniques and regression modeling for a more nuanced approach to depression detection.

Finally, ethical concerns, particularly in studies leveraging social media data for depression detection, are also a significant consideration, necessitating a delicate balance between innovation and privacy, as highlighted in  \cite{liu2022detecting} and \cite{yan2022challenges}.

In summary, the kaleidoscope of methodologies ranges from the traditional - decision trees and SVMs, explored in Yang, Jiang, and He \cite{yang2016decision} and Yang et al.  \cite{39yang2019depression}, to the avant-garde - deep learning architectures that carve out nuanced patterns within multimodal data sets, as highlighted in Squires et al. \cite{squires2023deep} and Krishna and Anju \cite{krishna2020different}. These methodologies are not without their challenges, as elucidated by Cummins et al. \cite{cummins2018speech}, where the limitations of dataset sizes and deep learning's voracious appetite for data become apparent.

Although these articles use the same dataset used in our work, we were able to spot several differences. Our work, utilizes the DAIC-WOZ database predominantly for PTSD patients, implementing sentiment analysis within NLP to analyze clinical interviews. In contrast, the approach of Liu et al.  \cite{liu2022detecting} capitalizes on the expanse of social media, engaging with a more volatile and expansive data collection environment. Our methodology provides a narrower scope compared to Liu et al.'s review of diverse social platforms for depressive symptom detection, emphasizing sentiment analysis tailored to PTSD.
In the realm of speech emotion analysis, while Bhavya et al. \cite{bhavya2022speech} narrows down to depression recognition via speech emotion analysis using the same database, our work extends the analysis to encompass a broader spectrum of mental disorders. This comprehensive approach allows us to evaluate classifier performance across different mental health conditions, including depression, anxiety, and PTSD.
Additionally, our work remains grounded in the domain of NLP tasks related to depression, utilizing ML classifiers, whereas Krishna and Anju \cite{krishna2020different} and Nguyen et al. \cite{nguyen2023multimodal} advocate for a multimodal approach. This highlights our specific focus on linguistic analysis as a powerful tool for mental health diagnostics, despite recognizing the value of multimodal approaches.

When addressing the challenges of dataset availability and ethical considerations, our work benefits from an established dataset, allowing us to sidestep broader ethical debates and focus on the optimization of classifiers. On the other hand, Silva et al. \cite{da2022review} and Yan et al. \cite{yan2022challenges} delve into these challenges, highlighting the ethical complexities inherent in AI for mental health, a perspective that enriches the dialogue within the field.

Furthermore, Aleem et al. \cite{aleem2022machine} and Ndaba et al. \cite{ndaba2023review} discuss class imbalance and feature engineering, proposing solutions like under-sampling techniques and regression modeling. While their contributions to the field are significant, our research primarily centers on the efficacy of various ML classifiers, showcasing the strengths of Random Forest and XGBoost.

Mao et al. \cite{mao2023systematic} provides a systematic review of automated clinical depression diagnosis, encompassing a broad range of studies and datasets. In contrast, our work offers a focused comparison of classifiers on a singular dataset, providing a detailed lens on classifier performance in PTSD-related depression detection.

Lastly, our work does not cover the advances presented by Nouman2022's exploration of contactless sensing technologies \cite{nouman2021recent}, nor does it delve into the deep learning for psychiatric applications as investigated by Squires et al. \cite{squires2023deep}. These studies present expanded views and forward-thinking perspectives on mental health monitoring and data analysis tools that differ from the direct application of specific classifiers evaluated in our work.

In this way, the most closely related works to our paper can be found among the articles utilizing advanced machine learning techniques and the DAIC-WOZ database for mental health diagnostics. Still, the work of Yang et al. \cite{39yang2019depression} employs methodologies akin to ours, such as Support Vector Machines (SVMs) and Text Convolutional Neural Network (TextCNN) for text-based emotional analysis, reflecting a similar application of sophisticated ML classifiers within the NLP framework. This parallel approach offers a complementary viewpoint to our work, enhancing the understanding of depression detection through linguistic cues and semantic analysis. The insights from \cite{39yang2019depression}, particularly the application of SVM and TextCNN, are invaluable for refining current practices and bolstering the predictive power of ML models in the domain of mental health, drawing a more comprehensive map of the current landscape and future directions in depression detection using machine learning and natural language processing.


\section{Experiment Design}

The objective of this case study is to build a diagnostic model for depression disorders based on different supervised ML models and NLP techniques. We explored multiple model tuning configurations, feature sets, and data preprocessing methodologies across all the models incorporated in our analysis. 

As a result, the design of this case study was conceived to identify the optimal model, the most suitable parameters, and any combination of factors that could provide the best results in terms of accuracy. We also established a baseline approach to check the efficiency of the models and examine the insights of our case study based on our results.

\begin{figure*}[!htb]
\centering
\includegraphics[scale=0.65]{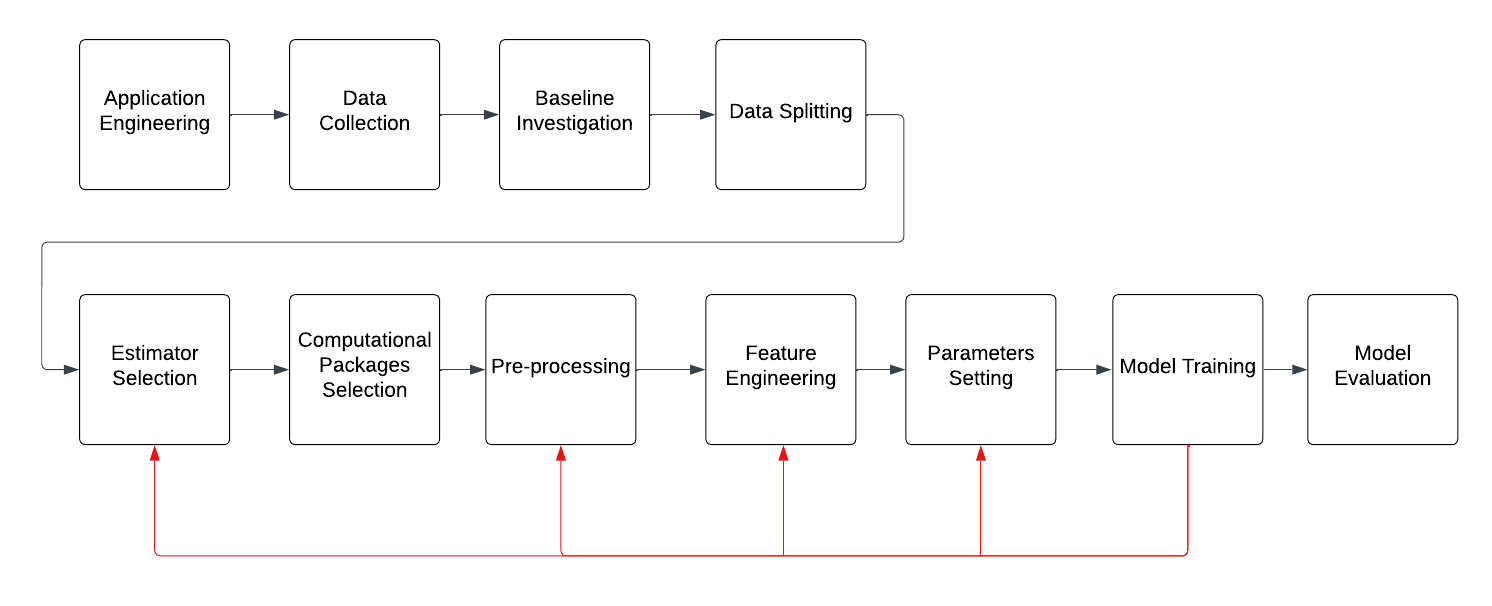}
\caption{Workflow Overview to select the optimum model. The red arrow illustrates that model training may loop back to estimator selection \cite{scikit-learn_settings}, pre-processing, feature engineering, and parameter setting.}
\label{fig:approach}
\end{figure*}

Figure \ref{fig:approach} showcases a comprehensive overview of the workflow we adopted to develop this model. This workflow is inspired by the one proposed by Amershi et al. \cite{amershi2019software}. Within this workflow, we delineated the steps that influenced the model selection. For instance, during the training phase, we explored two distinct preprocessing techniques, multiple feature combinations, varied parameters, and different selected estimators \cite{scikit-learn_settings} until we surpassed the baseline's performance. Subsequently, during the model evaluation phase, we assessed the final selected model using the safeguard dataset, which was previously separated during the data splitting phase.


As shown in the Figure \ref{fig:approach}, the final model selection is based on the chosen algorithm, 
a specific parameter configuration, and a set of features. Any changes in the algorithm, parameter tuning, or feature sets are regarded as distinct models (classifiers), each with its own results.

The next subsections describe the implementation details of each stage, tailored for this specific case study.

\subsection{Data Collection: Dataset}\label{AA}
We based our research on the Distress Analysis Interview Corpus - Wizard-of-Oz (DAIC-WOZ), a dataset designed to support the diagnosis of mental disorders such as depression, anxiety, and post-traumatic stress disorder. DAIC is a database that is part of a larger corpus \cite{10gratch2014distress} 
available to the research community by request. We have submitted a request and the data released refers only to the depressed patients' database.

This database contains clinical interviews conducted by humans, human-controlled agents, and autonomous agents. The computer agent is an animated virtual interviewer robot called Ellie that identifies mental illness indicators. Data includes 189 sessions of interviews which correspond to questionnaire responses and audio and video recordings. Each interview is identified by a session number and has a correspondent folder of files which includes files related to video features, audio, transcript, and audio features. This dataset contains training, development, and test subsets. Interviewees are both distressed and non-distressed individuals. 

We used the text files related to the transcript of the interviews. The Patient Health Questionnaire depression scale (PHQ-8) defines the depression diagnostic and severity measure.  and the PHQ-8 file which described the score of each patient according to the depression scale described in  \cite{11kroenke2009phq}.

\subsection {Baseline Investigation}
Before determining what techniques will be used in our case study we established a baseline approach. Because our classifier will choose among depressed and not depressed states (0 or 1), our approach consists in setting the same state (e.g. depressed - 0) to each observation of our training set and then checking the error rate in the predictions of our testing set. The purpose is to compare the accuracy rate of our chosen models not only amongst themselves but also with the baseline approach to gauge the model's efficacy.

\subsection{Data Splitting: Training and Testing Set}
For this study, we will use approximately 80\% of our data set as a training set (150 observations/ interviews/ transcripts) and the remaining approximately 20\% (38 observations/ interviews/ transcripts) as our testing set. 

Because three meetings were removed from the data set due to structural problems, we ended up with 148 interviews in our training set and 37 in our testing set.

\subsection{Estimator Selection}

As outlined in \cite{tavares2022adaptive}, certain heuristics enable the pre-selection of machine learning estimators \cite{scikit-learn_settings} that are best suited for a particular dataset or problem. By doing so, during the pursuit of the optimal model, it is possible to narrow down a group of most compatible estimators, thereby reducing our search options. Given the characteristics of the chosen problem and dataset (including its labeled nature, number of labels, size, number of features, and application output), and the algorithms that we identified in the related works (section \ref{sec:related}), our exploration was confined to three machine-learning techniques. They are:

\begin{itemize}
    
\item Random Forest Classifier (RF): an algorithm that builds multiple decision trees and makes a final prediction based on the majority outcome of each decision tree.
\item XGBoost (XGB): an algorithm that sequentially builds decision trees that recurrently diminish the error of the previous prediction by applying gradient descent on a loss function when adding new trees.
\item Support Vector Machine (SVM): an algorithm that maps the data points to another space (using kernel algorithms) so that it is possible to separate (and therefore classify) the data points into different categories. 
\end{itemize}

\subsection{Pre-processing: Data Cleaning}
Different attempts were made to clean the data. Initially, a custom set of stopwords was removed from the conversation, but the end result of the classification came out worse in this scenario as opposed to when no words were extracted. Moreover, the conversations had basic punctuation removed and the letters converted to lowercase. Afterward, the artificial conversation markers in the dataset (which did not represent words actually spoken in the dialogue, but rather indicated what kind of comment was being made) were removed to keep the text as close as possible to real conversations. Finally, the datasets that only contained interviewee speeches (and no comments / questions from the bot) were not used, given that most of the features in the model relied on questions asked by the bot.

\subsection{Feature Engineering: Features Selection}
Although the objective is to test a different set of features in our chosen NLP algorithm, it is important to mention that some of our features, such as the Sentiment Analysis calculation and the average of unique words or stop words, were also derived from other NLP techniques. 

We have built a total of 27 features (since the first feature described contains 19 separated features) which were combined in different ways and applied across the 3 NLP classifiers to find the best combination in terms of model accuracy. The list of the features considered on this study and its explanation can be found in Table \ref{table:tableI}. 
  
\begin{table*}[!htb]
    \centering
    \caption{Explanation of Features}
    \label{table:tableI}
    \begin{tabularx}{\textwidth}{lX}
        \toprule
        \textbf{Feature} & \textbf{Description} \\
        \midrule
        Answers sentiments as Features & The data set contains a transcription of a conversation between a bot and a person, during which the bot asks its counterpart several questions. The sentiment of the answers to the most relevant questions are used as features in our model. We have actually worked with 19 sentences/questions with this purpose. The most relevant sentences found in the database are displayed in the Results section. \\
        Sentiment of an Interview (avg sentiment) & Sentiment of all of the comments made by the person throughout the conversation with the bot. \\
        Average Response Time (avg response time) & Average time for interviewee to answer a question from the bot. \\
        Speech speed & Calculated by measuring the number of words for each comment of the interviewee, measuring the time it took to make that comment, dividing those two values, and taking the average of that result for all comments made by the interviewee throughout the conversation. \\
        Average Unique Words Frequency (Avg unique frequency) & Calculated by measuring the number of unique (without counting repetition) words spoken in a comment, then measuring the number of words in that comment, dividing the first by the second, and taking the average of that calculation for all comments made by the interviewee throughout the conversation. \\
        Average Stop Words Frequency (Avg sw frequency) & Calculated similarly to the previous metric (avg unique frequency), but replacing unique words with stop words, which are words without expressive content (e.g. uh, um, mm). The definition of what constitutes a stop word is flexible, but there are implementations from libraries such as NLTK that contain a set of stop words. Our set is an extension of NLTK’s set of stop words. \\
        Average Number of Characters (Avg characters) & Calculated by measuring the number of characters of each comment, dividing that value by the number of words in that comment, and then averaging that calculation through all comments made by the interviewee in the meeting. \\
        Average Noun, Verbs, Adjectives, Adverbs (Avg nouns, avg verbs, adj freq, avg adv) & Calculated similarly to the stop word frequency (Avg sw frequency), but instead of stop words, it calculates the average frequency of nouns, verbs, adjectives and adverbs, respectively. \\
        Average of First-Person Related Words (fp avg) & Similar to stop word frequency (Avg sw frequency), but instead of stop words, it measures the frequency of first-person-related words (e.g. "I", "we", "us"). \\
        \bottomrule
    \end{tabularx}
\end{table*}

Finally, it is also worth mentioning that the best results from our case study were obtained after testing several combinations of features from this list (and even features that were originally considered but then discarded because of poor results) including the sentiment of all the interviews and the sentiment of the answers from some selected questions that appeared in most of the transcripts.  
Additional information about the features selection can be found in the Results section.

\subsection{Parameters Setting}

\paragraph{Random Forest Classifier}
This classifier has 18 parameters, out of which max depth and random state were set to non-default values. Random state is used to promote the reproducibility of the model, whereas max depth influences overfitting tendencies, which is why it was included in the training process.

\paragraph{XGBoost}
This classifier has 33 parameters, out of which n estimators, eval metric, learning rate, label encoder, max depth, and random state were set to non-default values. Random state is used to promote the reproducibility of the model, whereas the max depth influences the overfitting tendency and is thus included in the training process. 27 of the parameters for this model are optional.

\paragraph{Support Vector Machine}
This classifier has 15 parameters, out of which the kernel, gamma, and random state were set to non-default values. Random state is used to promote the reproducibility of the model, whereas the max depth influences overfitting tendency and is thus included in the training process.

\subsection{Computational Packages Selection}


We have used the Scikit-learn library for implementing the Random Forest Classifier, the XGBoost, and the SVM algorithms. Scikit-learn is one of the most versatile and reliable library for machine learning in Python, as it features several useful tools for machine learning and statistical modeling, such as classification, regression, clustering, and dimensionality reduction via a consistency interface in Python.

Some of the features used in this study were calculated by using the following packages:

\begin{itemize}
  
 \item 	For Sentiment Analysis calculation (of a sentence or an interview): TextBlob library, which is a Python (2 and 3) library for processing textual data. It provides a simple API for diving into common natural language processing (NLP) tasks such as part-of-speech tagging, noun phrase extraction, sentiment analysis, classification, translation, and more.

\end{itemize}

\begin{itemize}
   
 \item 	For Features Extraction (Average Unique Words Frequency, Average Stop Words Frequency): NLTK (Natural Language Processing Toolkit, which is a leading platform for building Python programs that work with human language data. It provides easy-to-use interfaces for over 50 corpus and lexical resources such as WordNet, along with a suite of text processing libraries for classification, tokenization, stemming, tagging, parsing, semantic reasoning and wrappers for industrial-strength NLP libraries.

\end{itemize}

\section{Results}

As mentioned in the Experiment Design section, for this Case Study we've tested some of the most adopted machine learning models in the scientific literature with different sets of features and sets of parameters to establish how reliable are these models in the detection of depression disorders. 
In this sense we have adopted 3 NLP Classifiers with different set of features that were based on sentiment analysis and other natural language processing techniques. 

Our results suggest that Random Forest and XGBoost are the best classifiers for the task since they have presented the same accuracy level (84\%).
Although the SVM algorithm (Support Vector Classifier) have returned poor results in comparison with the other two models, we were able to replicate the results found in the comparable scientific literature. 
All the tested models results in accuracy levels beyond our baseline approach.
More details about the result for each model such as features selection and parameters tuning can be found in the following sub-sections.

\subsection{Baseline Approach Results}
After arbitrarily setting our data set with a prediction value of 1, we verified the error rate (in this case the accuracy level) of approximately 65\%. 

\subsection{Random Forest Model Results}

In our model classification process, we started with an initial group of 17 features. For each classification, we selected 4 distinct features from this group. With parameters kept fixed for the classifier, this approach resulted in the creation of 2,380 unique models. From these, the top-performing models achieved a notable accuracy of approximately 83.8\%. A comprehensive summary of these findings can be found in Tables \ref{tab:rf_details} and \ref{tab:top_features}.

\begin{table}[!htb]
\centering
\caption{Random Forest Model Details}
\label{tab:rf_details}
\begin{tabular}{|l|p{5cm}|}
\hline
Aspect & Description \\
\hline
Features & Initial group of 17, 4 chosen for each classification \\
\hline
Parameters & Fixed \\
\hline
Number of Models & 2,380 \\
\hline
Best Accuracy & 83.8\% \\
\hline
\end{tabular}
\end{table}

\subsection{XGBoost Model Results}

During the model training process, we experimented with varying numbers of estimators. Specifically, the "n estimators" took on values of 100, 300, and 500. By selecting 4 out of the 17 available features, we generated 2,380 unique feature sets. These were then paired with each of the 3 estimator configurations, resulting in a total of 7,140 classifiers for our model's training. Notably, the peak accuracy achieved remained consistent at around 83.8\%. For a comprehensive breakdown of the features and parameters yielding the best outcomes, refer to Tables \ref{tab:xgb_details} and \ref{tab:top_features_xgb}.

\begin{table}[!htb]
\centering
\caption{XGBoost Model Details}
\label{tab:xgb_details}
\begin{tabular}{|l|p{5cm}|}
\hline
Aspect & Description \\
\hline
Features & Initial group of 17, 4 chosen for each classification \\
\hline
Parameters & Different ``n estimators": 100, 300, 500 \\
\hline
Number of Models & 7,140 \\
\hline
Best Accuracy & 83.8\% \\
\hline
\end{tabular}
\end{table}

\subsection{SVM Model Results}

Initially, we categorized our data into two distinct groups: the first encompassing 17 features and the second, 19. From these, we handpicked 4 features from each group, which culminated in 2,380 and 11,628 unique feature sets for the first and second groups, respectively. For our SVM model, we focused on optimizing two parameters: Kernel and Gamma. Our testing involved 5 variations of Gamma (1, 0.1, 0.01, 0.001, and auto) paired with 2 Kernel options ('rbf' and 'linear'). As a result of this comprehensive approach, we devised 23,800 SVM models from the first feature group and 116,280 from the second. By synergizing every feature set with the Gamma and Kernel variations, the total models per group remained consistent. Among all models tested, the pinnacle of accuracy achieved was approximately 64.8\%. The detailed configurations of the model that delivered the best performance can be found in Tables \ref{tab:svm_details} and \ref{tab:top_features_svm}.


\begin{table}[!htb]
\centering
\caption{SVM Model Details}
\label{tab:svm_details}
\begin{tabular}{|l|p{5cm}|}
\hline
Aspect & Description \\
\hline
Feature Groups & Group 1: 17 features, Group 2: 19 features \\
\hline
Feature Combinations & Group 1: 2,380 sets, Group 2: 11,628 sets \\
\hline
Parameters & Kernell: 'ref', 'linear'; Gamma: 1, 0.1, 0.01, 0.001, auto \\
\hline
Number of Models & Group 1: 23,800, Group 2: 116,280 \\
\hline
Best Accuracy & 64.8\% \\
\hline
\end{tabular}
\end{table}

\subsubsection{Increasing the Number of Features}

In a second attempt, classifications were made using over 5000 combinations of 10 different features chosen from a set of 20 features. Only the "rbf" kernel algorithm was used in this case. The 'gamma' parameter assumed values 1 and 'auto', while the "C" parameter assumed values 1,5, and 10.

Thus, we have 184,756 possible sets of features from the choice of 10 features from a set of 20 features.
By combining each set of features with the 3 options of C parameter, 1 option of Kernel, and 2 options for Gamma, we would ended up with 1,108,536 classifiers for training our model. We tested approximately 30,000 of these classifiers for training our model. The 5 classifiers representing the best results obtained are shown below:

\begin{table}[!htb]
\centering
\caption{Classifier Configuration Details}
\label{tab:classifier_details}
\begin{tabular}{|l|p{4cm}|}
\hline
Aspect & Description \\
\hline
Total Features & 20 \\
\hline
Feature Combinations & Over 5000 combinations of 10 different features \\
\hline
Kernel & rbf \\
\hline
Gamma & 1, auto \\
\hline
C Parameter & 1, 5, 10 \\
\hline
Total Possible Classifiers & 1,108,536 \\
\hline
Classifiers Used & Approximately 30,000 \\
\hline
Best Accuracy & 70.3\% \\
\hline
\end{tabular}
\end{table}

\section {Discussion}
\subsection{General Insights from the Case Study}
Our results are not only better than the baseline approach in terms of accuracy but also outperformed the results obtained in the scientific literature. As a benchmark for this study we used \cite{39yang2019depression}
which is an article that tested different NLP techniques for Depression detection based on the same data set used in our study. 

More specifically, the Random Forest Model and XGBoost Model results represented a significant improvement in terms of accuracy (approximately 84\%) in comparison with the models used  by the authors in \cite{39yang2019depression} (SVM and CNN with approximately 70\% and 72\% of accuracy, respectively).





    

Regarding the data cleaning e preprocessing in our case study we made text lower case and removed basic punctuation (e.g., ",", ".", "[", "]"), and also all the text content that was not actually present in the conversation, but rather indicated what kind of comment was being made by the bot. Moreover, 3 meetings were removed from the dataset due to structural problems (did not have comments –any or most – from the bot). Additionally, because 
the sentiment of the relevant questions had a central role among our selected features, we have realized that there was no group of questions (or single question) included in every single transcript (interview). Thus, we have used the questions that could be found in most of the interviews. In this sense,  questions that were not in all meetings were assumed to yield an answer of sentiment 0 for each of the meetings in which they did not appear.

\subsection{Insights from the Sentiment Features in Top-Performing Models}
When examining the top-performing models, RF and XGB, we observed a recurring prominence of features such as Average Response Time, Speech Speed, and Average Number of Characters. Intriguingly, within each group of four features, only one was tied to a sentiment score. However, this score wasn't reflective of the overall meeting sentiment but was specific to an answer to a certain question. Notably, the particular questions associated with this sentiment score varied across the best-performing models.

This leads us to several insights regarding the sentiment feature:

\begin{itemize}

    \item The sentiment feature, linked to the response of a specific question, consistently occupies just one slot in every four-feature group.
    \item The exact question associated with the sentiment feature isn't uniform across the top models.
\end{itemize}

From this, we can deduce that the sentiment score, specific to an answer, may not consistently influence the model's outcomes. This inconsistency hints that the sentiment attached to a particular question's response might not consistently sway the model's effectiveness. Yet, the recurring appearance of some form of sentiment feature in top models indicates that sentiment does play a part, though its influence may not be uniform. This irregularity could point towards potential inaccuracies in how sentiment is gauged. If a sentiment score tied to a specific question was a pivotal determinant, we would expect its consistent presence across all top models. To ascertain the relevance of the sentiment feature to the classifier's results, a comprehensive feature importance analysis would be essential.

\subsection{Insights from the Interplay Between Dataset Bias and Imbalance in PTSD-focused Studies}

In our study centered on patients with PTSD, there's an inherent assumption that the dataset may lean towards a positive depression diagnosis, owing to the known comorbidity between PTSD and depression. However, it's noteworthy that our dataset was imbalanced, with only 56 of the 188 interviews being from individuals diagnosed with depression. This poses an intriguing question: might the imbalanced nature of the dataset counterbalance or mitigate the anticipated bias introduced by focusing on PTSD? This potential balancing effect is one reason we primarily relied on accuracy as our key evaluation metric in this study. Still, a deeper exploration and analysis are needed to fully understand the interplay between dataset bias and imbalance in this context.

\section{Threats to Validity} 
Although there are some issues related to the small dataset used in our study and the fact that we used a single evaluation metric, namely model accuracy, the biggest threat to validity of our results stems from the generalization issues related to the inner characteristics of our dataset. In addition, the dataset is not well distributed because most of the participants were not classified as depressed at the time of the interviews.



\begin{table*}[!htb]
\centering
\caption{Random Forrest - Top Model Features with Best Results for Random Forrest}
\label{tab:top_features}
\begin{tabular}{|c|p{9cm}|c|}
\hline
Rank & Features & Accuracy \\
\hline
1 & avg response time, avg nouns, adj freq, what's your dream job & 83.8\% \\
\hline
2 & avg response time, avg nouns, adj freq, do you consider yourself an introvert & 83.8\% \\
\hline
3 & avg response time, avg nouns, adj freq, what do you do to relax & 83.8\% \\
\hline
4 & avg response time, avg nouns, adj freq, how are you at controlling your temper & 83.8\% \\
\hline
5 & avg response time, avg nouns, adj freq, last time you argued and its reason & 83.8\% \\
\hline
\end{tabular}
\end{table*}


\begin{table*}[!htb]
\centering
\caption{XGBoost - Top Model Features and Parameters with Best Results for XGBoost}
\label{tab:top_features_xgb}
\begin{tabular}{|c|p{6cm}|p{4cm}|c|}
\hline
Rank & Features & Parameters & Accuracy \\
\hline
1 & avg response time, speech speed, avg nouns, controlling temper & use\_label\_encoder: False, eval metric: auc, learning rate: 0.05, max depth: 12, n estimators: 300 & 83.8\% \\
\hline
2 & avg response time, speech speed, avg nouns, close to family & use\_label\_encoder: False, eval metric: auc, learning rate: 0.05, max depth: 12, n estimators: 300 & 83.8\% \\
\hline
3 & avg response time, speech speed, avg nouns, any regrets & use\_label\_encoder: False, eval metric: auc, learning rate: 0.05, max depth: 12, n estimators: 300 & 83.8\% \\
\hline
4 & avg response time, speech speed, avg nouns, memorable experiences & use\_label\_encoder: False, eval metric: auc, learning rate: 0.05, max depth: 12, n estimators: 300 & 83.8\% \\
\hline
\end{tabular}
\end{table*}

\begin{table*}[!htb]
\centering
\caption{SVM - Top Model Features and Parameters with Best Results for SVM}
\label{tab:top_features_svm}
\begin{tabular}{|c|p{6cm}|p{4cm}|c|}
\hline
Rank & Features & Parameters & Accuracy \\
\hline
1 & avg nouns, avg response time, speech speed, avg characters & kernel: rbf, gamma: auto & 64.8\% \\
\hline
2 & how do you like your living situation, avg response time, speech speed, avg characters & kernel: rbf, gamma: auto & 64.8\% \\
\hline
3 & how do you like your living situation, when you don't sleep well, avg response time, speech speed, avg characters & kernel: linear, gamma: auto & 64.8\% \\
\hline
4 & do you feel down, avg response time, speech speed, avg characters & kernel: rbf, gamma: auto & 64.8\% \\
\hline
5 & when last felt happy, avg response time, speech speed, avg characters & kernel: rbf, gamma: auto & 64.8\% \\
\hline
\end{tabular}
\end{table*}


\begin{table*}[!htb]
\centering
\caption{Top 5 Classifier Features, Parameters, and Accuracies for SVM}
\label{tab:top_classifiers}
\begin{tabular}{|c|p{6cm}|p{4cm}|c|}
\hline
Rank & Sampled Features & Parameters & Accuracy \\
\hline
1 & avg response time, speech speed, avg characters, avg nouns, ... & kernel: rbf, gamma: 1, C: 10 & 70.3\% \\
\hline
2 & avg response time, speech speed, avg characters, avg nouns, ... & kernel: rbf, gamma: 1, C: 10 & 70.3\% \\
\hline
3 & avg response time, speech speed, avg characters, avg nouns, ... & kernel: rbf, gamma: 1, C: 10 & 70.3\% \\
\hline
4 & avg response time, speech speed, avg characters, avg nouns, ... & kernel: rbf, gamma: 1, C: 10 & 70.3\% \\
\hline
5 & avg response time, speech speed, avg characters, avg nouns, ... & kernel: rbf, gamma: 1, C: 10 & 70.3\% \\
\hline
\end{tabular}
\end{table*}




In summary, the most sensitive issues regarding the studies of our case study are related to the possible lack of generalization of our results due to the inner characteristics of our dataset. However, the results are promising when we consider the assessment related to several alternatives involving models, data cleaning and pre-processing, selected features, and parameters.

\section{Conclusions and Future Work} 
\subsection{Conclusions}
This initial study focuses on developing a Depression Detection Model using Sentiment Analysis and other NLP techniques specifically for patients diagnosed with PTSD. In our research, we compared various classifiers, feature sets, and other factors to determine the most effective approach.  

We have trained our models with a dataset built from the transcripts from 188 sessions of clinical interviews which correspond to questionnaire responses and audio and video recordings. The interviews were conducted by autonomous agents (bot). Our approach enabled us to achieve better results in terms of accuracy then we had in the comparable literature based on the same data set. The main difference between our case study and the comparable literature relied precisely on the model choices, features selection and data cleaning and processing of our data set. 

After determining the best techniques, models and features, we tested thousands (sometimes tens of thousands) of combinations of features and parameters to achieve optimal results for each model and technique. This exhaustive analysis not only enhanced the accuracy and reliability of our chosen models but also provided a deeper understanding of the intricate relationships between various features and parameters. This insight paves the way for more targeted and efficient implementations in subsequent studies.


In summary, we identified compelling evidence indicating numerous opportunities for advancement in the future development of a depression diagnostic system utilizing NLP techniques. These potential improvements are multifaceted. They encompass the exploration of a wider range of model choices, rigorous data cleaning and pre-rocessing methodologies, and, most prominently, refined strategies in feature selection and intricate feature engineering

\subsection{Future Work}

The findings of our study chart the way for several avenues of future research, primarily centered around expanding our methodology, refining feature selection, enhancing data preprocessing, and bolstering model generalization.

In terms of model selection, our comparative analysis could be broadened to include other ML classifiers such as Decision Trees, Convolutional Neural Networks (CNN), and BERT (Bidirectional Encoder Representations from Transformers). Given the state-of-the-art performance of BERT and its derivatives (e.g., RoBERTa, DistilBERT) across various NLP tasks, we propose exploring CNN and BERT-based models for feature generation, particularly by utilizing vector space representations for textual data. Further optimization studies for kernel algorithms might enhance the performance of the SVM model. Moreover, adopting Transformer models like BERT and leveraging Large Language Models (LLM) can refine sentiment score calculations, enhancing our classifiers' feature selection. This progression should culminate in a detailed feature importance analysis.

In terms of feature selection enhancement, lexicon-based approaches using dictionaries specific to medical conditions can be invaluable. Generating features such as word frequency, sentiment, and more can be beneficial. Additionally, integrating TF-IDF for pinpointing significant transcript words and incorporating bigrams can enrich contextual understanding and sentiment score assignment. An intriguing perspective we encountered centers on assigning variable weights to negative words based on context, achieved through combining lexicon dictionaries and context-weighted sentiment analysis.

Exploring new features such as the frequency of the word “I”, use of possessive pronouns, past tense verb occurrences, and readability scores can add depth to our analysis. We also advocate for refined preprocessing methods, such as leveraging the original NLTK set for stop word removal and incorporating lemmatization, a staple in NLP and machine learning.

Addressing dataset bias and imbalance remains pivotal. Potential research angles include assessing if the dataset imbalance can mitigate biases introduced by emphasizing PTSD sufferers, given their established association with depression. It is also relevant to implement rebalancing techniques that can enhance model generalization, and explore larger, more balanced datasets to prevent overfitting and ensure robust model generalization.

\bibliographystyle{IEEEtran}
\bibliography{bda4hm}

\begin{thebibliography}{10}
\providecommand{\url}[1]{#1}
\csname url@samestyle\endcsname
\providecommand{\newblock}{\relax}
\providecommand{\bibinfo}[2]{#2}
\providecommand{\BIBentrySTDinterwordspacing}{\spaceskip=0pt\relax}
\providecommand{\BIBentryALTinterwordstretchfactor}{4}
\providecommand{\BIBentryALTinterwordspacing}{\spaceskip=\fontdimen2\font plus
\BIBentryALTinterwordstretchfactor\fontdimen3\font minus \fontdimen4\font\relax}
\providecommand{\BIBforeignlanguage}[2]{{%
\expandafter\ifx\csname l@#1\endcsname\relax
\typeout{** WARNING: IEEEtran.bst: No hyphenation pattern has been}%
\typeout{** loaded for the language `#1'. Using the pattern for}%
\typeout{** the default language instead.}%
\else
\language=\csname l@#1\endcsname
\fi
#2}}
\providecommand{\BIBdecl}{\relax}
\BIBdecl

\bibitem{3saqib2021machine}
K.~Saqib, A.~F. Khan, and Z.~A. Butt, ``Machine learning methods for predicting postpartum depression: scoping review,'' \emph{JMIR mental health}, vol.~8, no.~11, p. e29838, 2021.

\bibitem{4lee2018applications}
Y.~Lee, R.-M. Ragguett, R.~B. Mansur, J.~J. Boutilier, J.~D. Rosenblat, A.~Trevizol, E.~Brietzke, K.~Lin, Z.~Pan, M.~Subramaniapillai \emph{et~al.}, ``Applications of machine learning algorithms to predict therapeutic outcomes in depression: A meta-analysis and systematic review,'' \emph{Journal of affective disorders}, vol. 241, pp. 519--532, 2018.

\bibitem{5spasic2020clinical}
I.~Spasic, G.~Nenadic \emph{et~al.}, ``Clinical text data in machine learning: systematic review,'' \emph{JMIR medical informatics}, vol.~8, no.~3, p. e17984, 2020.

\bibitem{12ganguli2020survey}
R.~Ganguli, A.~Mehta, and S.~Sen, ``A survey on machine learning methodologies in social network analysis,'' in \emph{2020 8th International Conference on Reliability, Infocom Technologies and Optimization (Trends and Future Directions)(ICRITO)}.\hskip 1em plus 0.5em minus 0.4em\relax IEEE, 2020, pp. 484--489.

\bibitem{13malviya2021transformers}
K.~Malviya, B.~Roy, and S.~Saritha, ``A transformers approach to detect depression in social media,'' in \emph{2021 International Conference on Artificial Intelligence and Smart Systems (ICAIS)}.\hskip 1em plus 0.5em minus 0.4em\relax IEEE, 2021, pp. 718--723.

\bibitem{14nandanwar2021depression}
H.~Nandanwar and S.~Nallamolu, ``Depression prediction on twitter using machine learning algorithms,'' in \emph{2021 2nd Global Conference for Advancement in Technology (GCAT)}.\hskip 1em plus 0.5em minus 0.4em\relax IEEE, 2021, pp. 1--7.

\bibitem{15yuki2020detecting}
J.~Q. Yuki, M.~M.~Q. Sakib, Z.~Zamal, S.~H. Efel, and M.~A. Khan, ``Detecting depression from human conversations,'' in \emph{Proceedings of the 8th International Conference on Computer and Communications Management}, 2020, pp. 14--18.

\bibitem{16tadesse2019detection}
M.~M. Tadesse, H.~Lin, B.~Xu, and L.~Yang, ``Detection of depression-related posts in reddit social media forum,'' \emph{Ieee Access}, vol.~7, pp. 44\,883--44\,893, 2019.

\bibitem{17azam2021identifying}
F.~Azam, M.~Agro, M.~Sami, M.~H. Abro, and A.~Dewani, ``Identifying depression among twitter users using sentiment analysis,'' in \emph{2021 International Conference on Artificial Intelligence (ICAI)}.\hskip 1em plus 0.5em minus 0.4em\relax IEEE, 2021, pp. 44--49.

\bibitem{18hossain2021social}
M.~T. Hossain, M.~A.~R. Talukder, and N.~Jahan, ``Social networking sites data analysis using nlp and ml to predict depression,'' in \emph{2021 12th International Conference on Computing Communication and Networking Technologies (ICCCNT)}.\hskip 1em plus 0.5em minus 0.4em\relax IEEE, 2021, pp. 1--5.

\bibitem{19skaik2020using}
R.~Skaik and D.~Inkpen, ``Using twitter social media for depression detection in the canadian population,'' in \emph{Proceedings of the 2020 3rd Artificial Intelligence and Cloud Computing Conference}, 2020, pp. 109--114.

\bibitem{20alghamdi2020predicting}
N.~S. Alghamdi, H.~A.~H. Mahmoud, A.~Abraham, S.~A. Alanazi, and L.~Garc{\'\i}a-Hern{\'a}ndez, ``Predicting depression symptoms in an arabic psychological forum,'' \emph{IEEE Access}, vol.~8, pp. 57\,317--57\,334, 2020.

\bibitem{21rastogi2018multi}
N.~Rastogi, F.~Keshtkar, and M.~S. Miah, ``A multi-modal human robot interaction framework based on cognitive behavioral therapy model,'' in \emph{Proceedings of the Workshop on Human-Habitat for Health (H3): Human-Habitat Multimodal Interaction for Promoting Health and Well-Being in the Internet of Things Era}, 2018, pp. 1--6.

\bibitem{22zhang2021depa}
P.~Zhang, M.~Wu, H.~Dinkel, and K.~Yu, ``Depa: Self-supervised audio embedding for depression detection,'' in \emph{Proceedings of the 29th ACM international conference on multimedia}, 2021, pp. 135--143.

\bibitem{23maniar2021depression}
S.~Maniar, K.~Patil, B.~Rao, and R.~Shankarmani, ``Depression detection from tweets along with clinical tests,'' in \emph{2021 International Conference on Intelligent Technologies (CONIT)}.\hskip 1em plus 0.5em minus 0.4em\relax IEEE, 2021, pp. 1--6.

\bibitem{24zogan2021depressionnet}
H.~Zogan, I.~Razzak, S.~Jameel, and G.~Xu, ``Depressionnet: A novel summarization boosted deep framework for depression detection on social media,'' \emph{arXiv preprint arXiv:2105.10878}, 2021.

\bibitem{25jain2020mental}
M.~P. Jain, S.~S. Dasmohapatra, and S.~Correia, ``Mental health state detection using open cv and sentimental analysis,'' in \emph{2020 3rd International Conference on Intelligent Sustainable Systems (ICISS)}.\hskip 1em plus 0.5em minus 0.4em\relax IEEE, 2020, pp. 465--470.

\bibitem{26trotzek2018utilizing}
M.~Trotzek, S.~Koitka, and C.~M. Friedrich, ``Utilizing neural networks and linguistic metadata for early detection of depression indications in text sequences,'' \emph{IEEE Transactions on Knowledge and Data Engineering}, vol.~32, no.~3, pp. 588--601, 2018.

\bibitem{27uddin2022deep}
M.~Z. Uddin, K.~K. Dysthe, A.~F{\o}lstad, and P.~B. Brandtzaeg, ``Deep learning for prediction of depressive symptoms in a large textual dataset,'' \emph{Neural Computing and Applications}, vol.~34, no.~1, pp. 721--744, 2022.

\bibitem{28wongkoblap2021deep}
A.~Wongkoblap, M.~A. Vadillo, V.~Curcin \emph{et~al.}, ``Deep learning with anaphora resolution for the detection of tweeters with depression: Algorithm development and validation study,'' \emph{JMIR Mental Health}, vol.~8, no.~8, p. e19824, 2021.

\bibitem{7alabdulkreem2021prediction}
E.~Alabdulkreem, ``Prediction of depressed arab women using their tweets,'' \emph{Journal of Decision Systems}, vol.~30, no. 2-3, pp. 102--117, 2021.

\bibitem{29wang2020multimodal}
Y.~Wang, Z.~Wang, C.~Li, Y.~Zhang, and H.~Wang, ``A multimodal feature fusion-based method for individual depression detection on sina weibo,'' in \emph{2020 IEEE 39th International Performance Computing and Communications Conference (IPCCC)}.\hskip 1em plus 0.5em minus 0.4em\relax IEEE, 2020, pp. 1--8.

\bibitem{30mulay2020automatic}
A.~Mulay, A.~Dhekne, R.~Wani, S.~Kadam, P.~Deshpande, and P.~Deshpande, ``Automatic depression level detection through visual input,'' in \emph{2020 Fourth World Conference on Smart Trends in Systems, Security and Sustainability (WorldS4)}.\hskip 1em plus 0.5em minus 0.4em\relax IEEE, 2020, pp. 19--22.

\bibitem{31shickel2020automatic}
B.~Shickel, S.~Siegel, M.~Heesacker, S.~Benton, and P.~Rashidi, ``Automatic detection and classification of cognitive distortions in mental health text,'' in \emph{2020 IEEE 20th International Conference on Bioinformatics and Bioengineering (BIBE)}.\hskip 1em plus 0.5em minus 0.4em\relax IEEE, 2020, pp. 275--280.

\bibitem{32sanyal2021study}
H.~Sanyal, S.~Shukla, and R.~Agrawal, ``Study of depression detection using deep learning,'' in \emph{2021 IEEE International Conference on Consumer Electronics (ICCE)}.\hskip 1em plus 0.5em minus 0.4em\relax IEEE, 2021, pp. 1--5.

\bibitem{33musleh2022twitter}
D.~A. Musleh, T.~A. Alkhales, R.~A. Almakki, S.~E. Alnajim, S.~K. Almarshad, R.~S. Alhasaniah, S.~S. Aljameel, and A.~A. Almuqhim, ``Twitter arabic sentiment analysis to detect depression using machine learning.'' \emph{Computers, Materials \& Continua}, vol.~71, no.~2, 2022.

\bibitem{34low2020natural}
D.~M. Low, L.~Rumker, T.~Talkar, J.~Torous, G.~Cecchi, and S.~S. Ghosh, ``Natural language processing reveals vulnerable mental health support groups and heightened health anxiety on reddit during covid-19: Observational study,'' \emph{Journal of medical Internet research}, vol.~22, no.~10, p. e22635, 2020.

\bibitem{35nguyen2014affective}
T.~Nguyen, D.~Phung, B.~Dao, S.~Venkatesh, and M.~Berk, ``Affective and content analysis of online depression communities,'' \emph{IEEE transactions on affective computing}, vol.~5, no.~3, pp. 217--226, 2014.

\bibitem{36zaghouani2018large}
W.~Zaghouani, ``A large-scale social media corpus for the detection of youth depression (project note),'' \emph{Procedia computer science}, vol. 142, pp. 347--351, 2018.

\bibitem{37jagtap2021use}
N.~Jagtap, H.~Shukla, V.~Shinde, S.~Desai, and V.~Kulkarni, ``Use of ensemble machine learning to detect depression in social media posts,'' in \emph{2021 Second International Conference on Electronics and Sustainable Communication Systems (ICESC)}.\hskip 1em plus 0.5em minus 0.4em\relax IEEE, 2021, pp. 1396--1400.

\bibitem{38hassan2020recognizing}
S.~B. Hassan, S.~B. Hassan, and U.~Zakia, ``Recognizing suicidal intent in depressed population using nlp: a pilot study,'' in \emph{2020 11th IEEE Annual Information Technology, Electronics and Mobile Communication Conference (IEMCON)}.\hskip 1em plus 0.5em minus 0.4em\relax IEEE, 2020, pp. 0121--0128.

\bibitem{40haque2020transformer}
F.~Haque, R.~U. Nur, S.~Al~Jahan, Z.~Mahmud, and F.~M. Shah, ``A transformer based approach to detect suicidal ideation using pre-trained language models,'' in \emph{2020 23rd international conference on computer and information technology (ICCIT)}.\hskip 1em plus 0.5em minus 0.4em\relax IEEE, 2020, pp. 1--5.

\bibitem{liu2022detecting}
D.~Liu, X.~L. Feng, F.~Ahmed, M.~Shahid, J.~Guo \emph{et~al.}, ``Detecting and measuring depression on social media using a machine learning approach: systematic review,'' \emph{JMIR Mental Health}, vol.~9, no.~3, p. e27244, 2022.

\bibitem{bhavya2022speech}
S.~Bhavya, R.~C. Dmello, A.~Nayak, and S.~S. Bangera, ``Speech emotion analysis using machine learning for depression recognition: a review,'' 2022, pre-print on EasyChair.

\bibitem{chaves2022automatic}
A.~Chaves, C.~Kesiku, and B.~Garcia-Zapirain, ``Automatic text summarization of biomedical text data: A systematic review,'' \emph{Information}, vol.~13, no.~8, p. 393, 2022.

\bibitem{squires2023deep}
M.~Squires, X.~Tao, S.~Elangovan, R.~Gururajan, X.~Zhou, U.~R. Acharya, and Y.~Li, ``Deep learning and machine learning in psychiatry: a survey of current progress in depression detection, diagnosis and treatment,'' \emph{Brain Informatics}, vol.~10, no.~1, pp. 1--19, 2023.

\bibitem{cummins2018speech}
N.~Cummins, A.~Baird, and B.~W. Schuller, ``Speech analysis for health: Current state-of-the-art and the increasing impact of deep learning,'' \emph{Methods}, vol. 151, pp. 41--54, 2018.

\bibitem{krishna2020different}
S.~Krishna and J.~Anju, ``Different approaches in depression analysis: A review,'' in \emph{2020 International Conference on Computational Performance Evaluation (ComPE)}.\hskip 1em plus 0.5em minus 0.4em\relax IEEE, 2020, pp. 407--414.

\bibitem{nguyen2023multimodal}
T.~T. Nguyen, V.~H.-Q. Pham, D.-T. Le, X.-S. Vu, F.~Deligianni, and H.~D. Nguyen, ``Multimodal machine learning for mental disorder detection: A scoping review,'' \emph{Procedia Computer Science}, vol. 225, pp. 1458--1467, 2023.

\bibitem{nouman2021recent}
M.~Nouman, S.~Y. Khoo, M.~P. Mahmud, and A.~Z. Kouzani, ``Recent advances in contactless sensing technologies for mental health monitoring,'' \emph{IEEE Internet of Things Journal}, vol.~9, no.~1, pp. 274--297, 2021.

\bibitem{da2022review}
A.~C. da~Silva, R.~C. Santana, T.~H. de~Lima, M.~L. Teodoro, M.~A. Song, L.~E. Z{\'a}rate, and C.~N. Nobre, ``A review of the main factors, computational methods, and databases used in depression studies.'' \emph{HEALTHINF}, pp. 413--420, 2022.

\bibitem{yan2022challenges}
W.-J. Yan, Q.-N. Ruan, and K.~Jiang, ``Challenges for artificial intelligence in recognizing mental disorders,'' \emph{Diagnostics}, vol.~13, no.~1, p.~2, 2022.

\bibitem{song2018human}
S.~Song, L.~Shen, and M.~Valstar, ``Human behaviour-based automatic depression analysis using hand-crafted statistics and deep learned spectral features,'' in \emph{2018 13th IEEE International Conference on Automatic Face \& Gesture Recognition (FG 2018)}.\hskip 1em plus 0.5em minus 0.4em\relax IEEE, 2018, pp. 158--165.

\bibitem{janardhan2022improving}
N.~Janardhan and N.~Kumaresh, ``Improving depression prediction accuracy using fisher score-based feature selection and dynamic ensemble selection approach based on acoustic features of speech,'' \emph{Traitement du Signal}, vol.~39, no.~1, p.~87, 2022.

\bibitem{mao2023systematic}
K.~Mao, Y.~Wu, and J.~Chen, ``A systematic review on automated clinical depression diagnosis,'' \emph{npj Mental Health Research}, vol.~2, no.~1, p.~20, 2023.

\bibitem{aleem2022machine}
S.~Aleem, N.~u. Huda, R.~Amin, S.~Khalid, S.~S. Alshamrani, and A.~Alshehri, ``Machine learning algorithms for depression: diagnosis, insights, and research directions,'' \emph{Electronics}, vol.~11, no.~7, p. 1111, 2022.

\bibitem{ndaba2023review}
S.~Ndaba, ``Review of class imbalance dataset handling techniques for depression prediction and detection,'' \emph{Available at SSRN 4387416}, 2023.

\bibitem{yang2016decision}
L.~Yang, D.~Jiang, L.~He, E.~Pei, M.~C. Oveneke, and H.~Sahli, ``Decision tree based depression classification from audio video and language information,'' in \emph{Proceedings of the 6th international workshop on audio/visual emotion challenge}, 2016, pp. 89--96.

\bibitem{39yang2019depression}
C.~Yang, X.~Lai, Z.~Hu, Y.~Liu, and P.~Shen, ``Depression tendency screening use text based emotional analysis technique,'' in \emph{Journal of Physics: Conference Series}, vol. 1237, no.~3.\hskip 1em plus 0.5em minus 0.4em\relax IOP Publishing, 2019, p. 032035.

\bibitem{amershi2019software}
S.~Amershi, A.~Begel, C.~Bird, R.~DeLine, H.~Gall, E.~Kamar, N.~Nagappan, B.~Nushi, and T.~Zimmermann, ``Software engineering for machine learning: A case study,'' in \emph{2019 IEEE/ACM 41st International Conference on Software Engineering: Software Engineering in Practice (ICSE-SEIP)}.\hskip 1em plus 0.5em minus 0.4em\relax IEEE, 2019, pp. 291--300.

\bibitem{scikit-learn_settings}
S.-L. developers, ``Statistical learning: the setting and the estimator object in scikit-learn,'' p. Accessed: November 03 2023.

\bibitem{10gratch2014distress}
J.~Gratch, R.~Artstein, G.~M. Lucas, G.~Stratou, S.~Scherer, A.~Nazarian, R.~Wood, J.~Boberg, D.~DeVault, S.~Marsella \emph{et~al.}, ``The distress analysis interview corpus of human and computer interviews.'' in \emph{LREC}.\hskip 1em plus 0.5em minus 0.4em\relax Reykjavik, 2014, pp. 3123--3128.

\bibitem{11kroenke2009phq}
K.~Kroenke, T.~W. Strine, R.~L. Spitzer, J.~B. Williams, J.~T. Berry, and A.~H. Mokdad, ``The phq-8 as a measure of current depression in the general population,'' \emph{Journal of affective disorders}, vol. 114, no. 1-3, pp. 163--173, 2009.

\bibitem{tavares2022adaptive}
C.~Tavares, N.~Nascimento, P.~Alencar, and D.~Cowan, ``Adaptive method for machine learning model selection in data science projects,'' in \emph{2022 IEEE International Conference on Big Data (Big Data)}.\hskip 1em plus 0.5em minus 0.4em\relax IEEE, 2022, pp. 2682--2688.

\end{thebibliography}

 \end{document}